%% file: bmvc_final.tex
\documentclass{bmvc2k}

\title{Re-examining Distillation for Continual Object Detection}

\addauthor{Eli Verwimp}{eli.verwimp@kuleuven.be}{1}
\addauthor{Kuo Yang}{yang.kuo@huawei.com}{2}
\addauthor{Sarah Parisot}{sarah.parisot@huawei.com}{2}
\addauthor{Lanqing Hong}{honglanqing@huawei.com}{2}
\addauthor{Steven McDonagh}{steven.mcdonagh@huawei.com}{2}
\addauthor{Eduardo Pérez Pellitero}{e.perez.pellitero@huawei.com}{2}
\addauthor{Matthias De Lange}{matthias.delange@kuleuven.be}{1}
\addauthor{Tinne Tuytelaars}{tinne.tuytelaars@kuleuven.be}{1}

\addinstitution{KU Leuven}
\addinstitution{Huawei Noah's Ark Lab}

\usepackage{graphicx}

\usepackage{comment}
\usepackage{amsmath,amssymb} 

\usepackage{subcaption}
\usepackage{booktabs}
\usepackage{multirow}

\runninghead{Verwimp \etal}{Distillation for Continual Object Detection}

\def\eg{\emph{e.g}\bmvaOneDot}

\def\etal{\emph{et al}\bmvaOneDot}

\begin{document}

\maketitle

\begin{abstract}
Training models continually to detect and classify objects, from new classes and new domains, remains an open problem. In this work, we conduct an analysis of why and how object detection models forget catastrophically. We focus on distillation-based approaches in two-stage networks; the most-common strategy employed in contemporary continual object detection work. Distillation aims to transfer the knowledge of a model trained on previous tasks -the teacher- to a new model -the student- while it learns the new task. We show forgetting happens mostly in the classification head, where wrong, yet overly confident teacher predictions prevent student models from effective learning. Our analysis provides insights in the effects of using  distillation techniques, and serves as a foundation that allows us to propose improvements for existing techniques by detecting incorrect teacher predictions, based on current ground-truth labels, and by employing an adaptive Huber loss as opposed to the mean squared error for the distillation loss in the classification heads. 
\end{abstract}

\section{Introduction}
\label{sec:intro}

Continual learning (CL) is a fast growing research topic within machine learning research. Numerous methods have been proposed to make models learn from non-stationary streams of data and prevent catastrophic forgetting of older knowledge~\cite{li2017learning,rebuffi2017icarl,lopez2017gradient}. The challenges of continual learning can be understood in light of the stability-plasticity dilemma~\cite{mermillod2013stability}, observed in both artificial and biological neural networks. We seek to balance the stability of trained models (preserving knowledge acquired during training), with their ability to adapt to new data (plasticity). A crucial challenge is the prevention of catastrophic forgetting~\cite{french1999catastrophic}. This phenomenon is observed in neural networks whenever there is change in the input data distribution, causing the network to entirely forget how to classify or detect previous data. Designing models and architectures that achieve both stability and plasticity properties remains a difficult task, and often a trade-off has to be made.

\paragraph{}
Continual learning has been mainly studied in the context of image classification, considering both class-incremental settings~\cite{li2017learning,rebuffi2017icarl,lopez2017gradient}, where new classes are introduced over time, and domain-incremental settings~\cite{lenga2020continual}, where the input domain of a class shifts during training. Current popular techniques include revisiting a small set of old images, adding regularization losses to the classification loss and/or freezing parts of the model. Research involving more complex tasks, such as object detection, has been less common~\cite{peng2020faster,kj2021incremental,joseph2021towards}. While classification tasks usually aim to recognize a single object per image, detection brings additional challenges. Object detection solutions often combine classification with a class agnostic regression mechanism which has to accurately predict the location of an object, requiring to learn what is an object of interest and what belongs to the background. Detection models should detect all objects that are part of an image, without knowing beforehand whether there will be one, a dozen, or none. Continual object detection (COD) adds another challenge; objects learned during previous COD tasks can appear in subsequent tasks and vice versa, albeit without labels. This is in contrast to classification, where incremental tasks are fully disjoint. These extra challenges and requirements prevent current classification techniques from being applied directly to detection models. In light of these challenges, COD approaches most commonly rely on distillation-based regularization so as to transfer knowledge from a teacher model, trained on one task, to a student model to be trained on the subsequent task~\cite{peng2020faster,zhou2020lifelong,kj2021incremental,liu2020continual}.

\paragraph{}
Understanding why catastrophic forgetting occurs and how existing methods alleviate the issue has been investigated to some extent for classification tasks. Knoblauch \emph{et al.}~\cite{knoblauch2020optimal} 
approached the problem from a theoretical point of view, and Benzing~\cite{benzing2020understanding} provides an analysis and a unifying framework for regularization methods. Nonetheless, to the best of our knowledge, there exists no detailed analysis of COD, the mechanisms in which forgetting occurs, as well as the impact on distillation-type solutions. 

\paragraph{}
In this work, we seek to understand how object detection models forget catastrophically, and more specifically how distillation can alleviate forgetting. Our analysis shows that distillation allows effectively learning a region proposal network (RPN) that performs well on all tasks, but prevents the classification head to adapt properly to new classes, due to overly confident, erroneous, teacher predictions. To address this, we introduce two modifications to the distillation loss, which facilitate inclusion of new classes. Firstly, we rely on task specific bounding box annotations to identify teacher errors and adjust its influence; second, we replace the standard Mean Squared Error (MSE) with a Huber loss, affording more robust knowledge transfer between tasks. We show that our modifications to SOTA distillation based method Faster-ILOD provide consistent improvements, achieving state of the art performance in the class incremental setting.

\paragraph{}
\noindent To summarize, our contributions are the following: 
\begin{itemize}
    \item To the best of our knowledge, we are the first to propose a detailed analysis scheme of continual object detection models' performance, breaking down how individual detector components influence forgetting.\footnote{\url{https://github.com/VerwimpEli/Continual_Object_Detection}}
    \item We provide new insights on the behavior of distillation solutions and address their limitations using two simple modifications, and demonstrate their benefits on a standard distillation-based method and yield state of the art performance on class incremental benchmarks.
\end{itemize}

\section{Related Work}
\label{sec:related}

\paragraph{Continual Learning.} The taxonomy of Continual Learning benchmarks has become quite extensive over the last few years, we therefore discuss here only essential literature and refer the reader to \cite{delange2021continual,masana2020class} for a more extensive review. Usually, the distribution of the input data changes at discrete steps and \textit{task} is used to refer to all data between successive steps~\cite{van2019three,delange2021continual}. When a task contains new classes, this is called class-incremental learning~\cite{rebuffi2017icarl,masana2020class}. In domain-incremental learning, the distribution of already present classes changes instead~\cite{lenga2020continual}.
When only a single epoch is allowed on each task, this is the field of streaming continual learning~\cite{De_Lange_2021_ICCV,aljundi2019gradient,hayes2019memory}. For this work, we do allow more than one epoch per task. Note that this implies that the model is aware of when the input data changes. See \cite{van2019three} for an overview of the three scenarios mentioned above. Following \cite{delange2021continual}, existing continual learning methods can be split into three main categories. Regularization methods add extra losses to either keep a model close in parameter space to the final model of the previous task \cite{kirkpatrick2017overcoming,aljundi2018memory}, or encourage similar outputs for similar inputs \cite{li2017learning}. Replay methods store a small subset of the samples of previous tasks and reuse them during training of new tasks to prevent forgetting~\cite{chaudhry2019continual,Verwimp_2021_ICCV}. Parameter isolation methods~\cite{mallya2018packnet,serra2018overcoming} isolate parts of the network and dedicate them to old tasks, but this requires the model to know to which task an image belongs, which is not considered here. 


\paragraph{Continual  Object Detection} Knowledge distillation is a technique first proposed by Hinton \emph{et al.}~\cite{hinton2015distilling}, to transfer knowledge from a large (teacher) to a small (student) network. Li \emph{et al.}~\cite{li2017learning}, adapted the idea and applied it to a continual classification task. Later, Shmelkov \emph{et al.} \cite{shmelkov2017incremental} introduced the idea to COD, after which others took over the concept \cite{zhou2020lifelong,kj2021incremental,liu2020continual,peng2020faster}. Initially, Hinton \emph{et al.} used the Kullback–Leibler (KL) divergence of the output of student and teacher as the loss function. However, all previously mentioned works in COD use an $L_2$-loss (i.e. mean squared error or MSE), with the exception of \cite{zhou2020lifelong}, who use a smoothed $L_1$ for the regression outputs, and \cite{kj2021incremental}, who use the KL-divergence for the classification heads. These works mainly differ in which exact outputs they distill (\emph{e.g.} logits \emph{vs.} log-likelihood outputs) and some hyper-parameters (\emph{e.g.} ratio of distillation loss to other losses), but all rely predominantly on MSE losses. Besides distillation, rehearsal \cite{chaudhry2019continual} is a technique often used to reduce forgetting. Both \cite{joseph2021towards} and~\cite{kj2021incremental} use a replay memory to continually learn detection tasks. In the former, the memory is used to train an RPN~\cite{ren2015faster} to recognize unknown objects, while in the latter it is used to condition the gradient, as a form of meta-learning~\cite{flennerhag2019meta}. Acharya \emph{et al.} employ a basic form of rehearsal, but their focus is on efficient compression of past data~\cite{acharya2020rodeo}. 

\paragraph{} In this work we will focus on how distillation is and can be used to alleviate forgetting and leave the analysis of rehearsal in COD for future work. While one-stage networks have been considered ~\cite{li2019rilod,zhang2020class}, the majority of works use two-stage networks~\cite{shmelkov2017incremental,zhou2020lifelong,kj2021incremental,liu2020continual,joseph2021towards,acharya2020rodeo}. Therefore, we focus on two-stage networks, 
in part as this is a pervasive approach in current literature, but also as its granularity enables a relatively simpler analysis. More specifically, we use Faster-ILOD \cite{peng2020faster} as our prototypical distillation example. This Faster-RCNN~\cite{ren2015faster} based method is a state of the art approach relying solely on distillation. More recent methods build on Faster-ILOD \cite{kj2021incremental, liu2020continual, zhou2020lifelong}, using similar distillation losses, but additionally employing other techniques (\eg rehearsal). These make it more challenging to disentangle the effects of distillation and those of the other techniques, and therefore we base our analysis on Faster-ILOD.



\section{Analysis \& Method}
\label{sec:method}

In this section, we first introduce the COD setting and relevant evaluation protocols. We then elaborate on object detectors' failure modes, and how they can help us understand why and how forgetting occurs. Finally, we dig deeper into why MSE in the ROI-head distillation fails and propose improvements to cope with these issues. 

\subsection{Problem Formulation}
\label{sec:method:analysis}

A standard COD problem is defined as a sequence of object detection tasks, where each task is associated with a set of training images and corresponding bounding box annotations. For now, we focus our analysis on the class incremental COD setting, which is currently the most popular set-up on which methods are developed and evaluated. In this setting, each task comprises a set of unique classes for which annotations are available, effectively separating class annotations into disjoint subsets. An important nuance with respect to standard task incremental classification settings is that an image can be part of multiple tasks, while bounding box annotations cannot. Indeed, an image will be part of a task if it possesses at least one object from the classes associated with this task. Standard COD benchmarks often consist of two tasks, and are referred to as $n_1 + n_2$, where $n_1$ and $n_2$ are the number of classes in tasks $1$ and $2$ respectively. This setting proves complex enough to reveal real-world COD challenges, and yet remains a feasible test-bed for investigative model training and analysis. In the remainder of this section, we first discuss standard COD evaluation and its shortcoming, then detail our analysis process towards understanding how forgetting occurs in COD tasks. 
For the purpose of our analysis, we use the VOC2007 $10+10$ benchmark~\cite{pascal-voc-2007,shmelkov2017incremental}, consistent with a majority of previous work. This benchmark separates the highly popular PASCAL VOC 2007 object detection dataset into two tasks. The first task (T1) contains the first (alphabetically ordered) ten classes, and the second task (T2) the next ten.

\subsection{Disentangling Forgetting in COD}

We now detail our analysis process towards the role of distillation in two-stage COD methods and their shortcomings. Two-stage object detectors consist of three main components: a backbone that extracts semantic features, an RPN  proposing regions that likely contain an object and the ROI-head, responsible for classifying said regions. See \cite{ren2015faster} and \cite{girshick2015fast} for more details.
We rely on Faster-ILOD to evaluate the impact of distillation, so we briefly detail its workings here. Distillation losses are applied on the three main components of the object detector. The backbone is distilled using the MSE between the features of the teacher's backbone and those of the student. Likewise, for the RPN, an MSE-loss is used on \emph{all} proposals of both RPN's, both for the objectness scores and the regression outputs. To distill the ROI-heads, first 64 out of the 128 proposals with the highest objectness score proposed by the student's RPN are randomly selected, then they are evaluated both by the heads of the teacher and student, and used in a third MSE component. For further details, see \cite{peng2020faster}.


\subsubsection{Backbone and RPN}
\label{sec:backbone_rpn}
Initial experiments on VOC10+10 indicate that most of the forgetting happens in the ROI-head. Compared to the Faster-ILOD benchmark, freezing the backbone leads to a 1 mAP point decreases, and not employing distillation (i.e. fine-tuning) to a 0.1 mAP decrease, indicating that forgetting isn't crucially happening in the backbone. Using Faster-ILOD's MSE-distillation the RPN respectively finds $98.6$\% and $98.4$\% of the labeled objects at IOU-level 0.5, which shows that using distillation forgetting in the RPN can be mostly prevented. The effectiveness of distillation in these tasks is interesting observation, and hints at a fundamental difference between a classification task and a regression task such as bounding box detection. This is possibly explained by the domain-incremental nature of the regression task in contrast to the class-incremental classification task \cite{van2019three}. Given these observations, for the remainder of this paper, we continue with an analysis of the ROI-head and how its continual learning performance can be improved. See Supplementary for details on these preliminary experiments.

\subsubsection{ROI-Head}

The classification head can either classify candidate boxes as background, or assign one of the classes of interest. When learning a new task without a specific continual learning mechanism, the classification head will only see annotations of new classes, which results in a bias towards those classes. Due to the nature of the content found in the considered benchmark datasets; instances of classes, belonging to previous tasks (``old classes''), will be unannotated in new task images. Yet crucially, such instances may still be present in the samples of the new task (\emph{i.e.} their appearance is now marked as background). The crux of the problem involves classifying these old classes as background; rather than their actual label. As a result, a crucial task for COD models is to preserve identification of classes pertaining to previous tasks in the classification head. 

\begin{figure}[tb]
    \centering
    \includegraphics[width=1.0\linewidth]{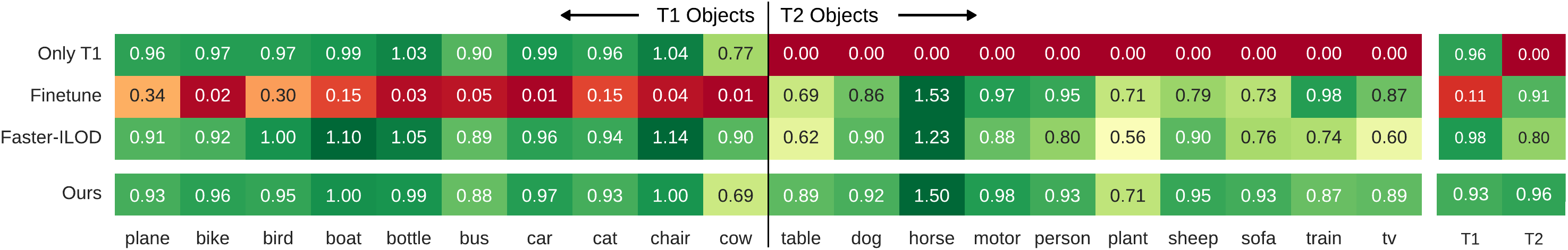}
    \vspace{0.05em}
    \caption{Normalized mAP of the classes in VOC after only learning T1, after finetuning on T2 and after learning T2 with Faster-ILOD and the final average per task. Ours is our improved version of Faster-ILOD, see Sections \ref{sec:improving_distillation} and \ref{sec:results} for further details. The results are normalized as $\frac{r_i}{r_{i, j}}$ with $r_i$ the mAP of class $i$ trained sequentially and $r_{i, j}$ the mAP of class $i$ of a model trained jointly on T1 and T2.}
    \label{fig:roi_analysis}
\end{figure}

\paragraph{}
Figure \ref{fig:roi_analysis} shows the mean average precision (mAP) of three models: one that's only trained on T1, one that's sequentially fine-tuned without any continual learning strategy on T2, and one that's trained as in Faster-ILOD. The results are normalized w.r.t. a joint model that is trained on all classes at once, which is commonly constituted as an upper bound in CL-settings. Unlike the RPN, the ROI-head forgets all but two classes catastrophically, indicating the need for continual learning techniques. The distillation used in Faster-ILOD is effective: the second task is learned well, while forgetting of the former task is prevented. 
Yet, for the classes of T2, the model using distillation only reaches an average of 80\% of the performance of a joint model. 
The individual results for the classes that make up the second task show that the errors aren't evenly distributed: some classes are learned as well or better than with a joint model, while classes like \textit{table}, \textit{plant} and \textit{tv} only reach about $60\%$.

\begin{figure}[h]
    \centering
    \includegraphics[width=0.60\linewidth]{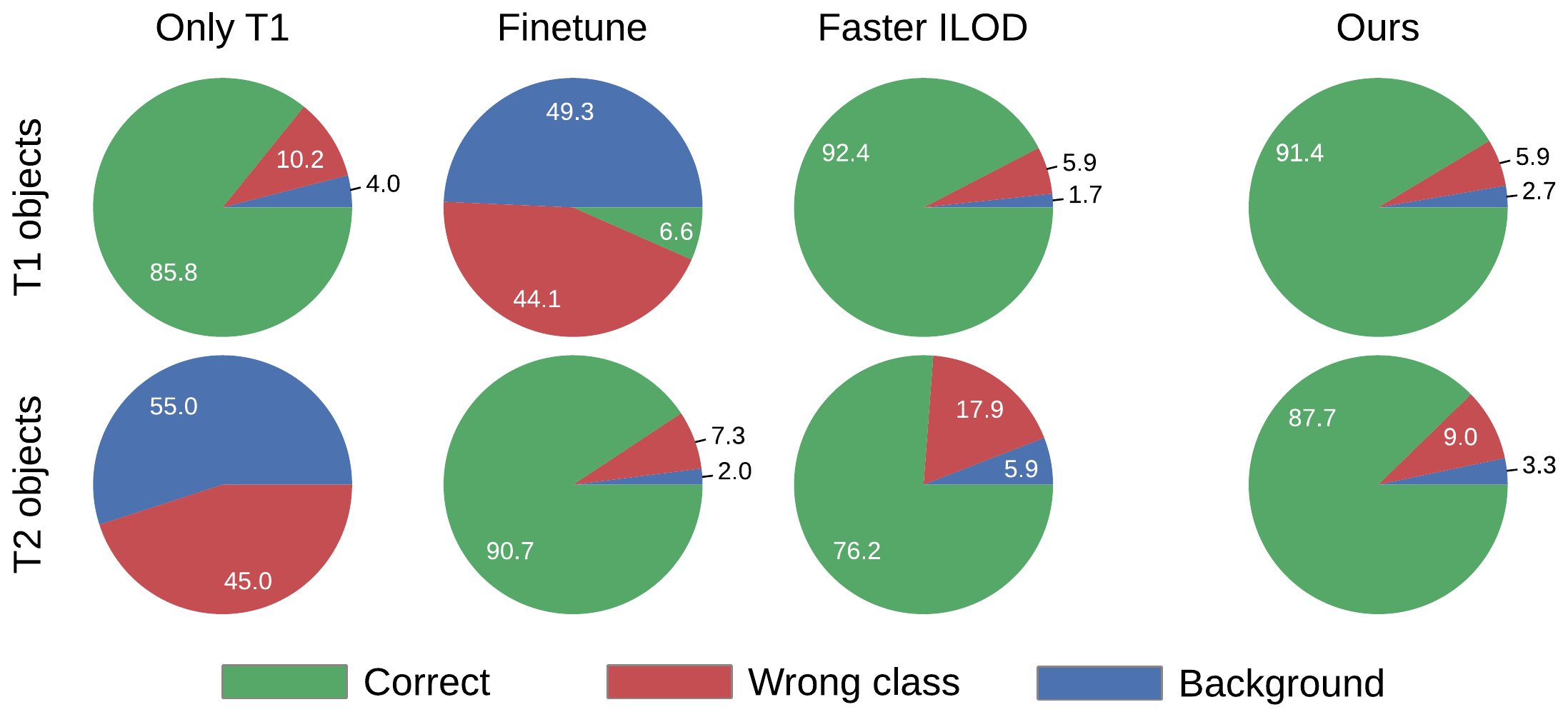}
    \vspace{0.4em}
    \caption{Partitioning of the classification of all region proposals, by a model trained on only the classes of T1, one that is subsequently also finetuned on T2, a Faster-ILOD model and our improvement, see sections \ref{sec:improving_distillation} and \ref{sec:results}.}
    \label{fig:roi_out}
\end{figure}

To improve upon the Faster-ILOD score, it is essential to understand why some of the classes are not properly learned. From all detected objects of T2, the Faster-ILOD model classifies $76.2\%$ correctly, assigns $17.9\%$ to the wrong class and marks $5.9\%$ as background, see Figure \ref{fig:roi_out}. While the number of proposals mistakenly classified as background is higher than in the first task, the wrongly classified samples make up the majority of the errors, hence our focus will be on those. Of all the wrongly classified proposals, 84\% are classified as a T1 class and only 16\% gets mixed up with another T2 class, see Supplementary for the full confusion matrices. This indicates that the model is overly static, which we conjecture is caused by the effects of the distillation loss. In the following section, we propose to relax this objective towards mitigating this major source of error.

\subsection{Improving ROI-Distillation}
\label{sec:improving_distillation}

\begin{figure}[b]
    \centering
    \includegraphics[width=0.65\linewidth]{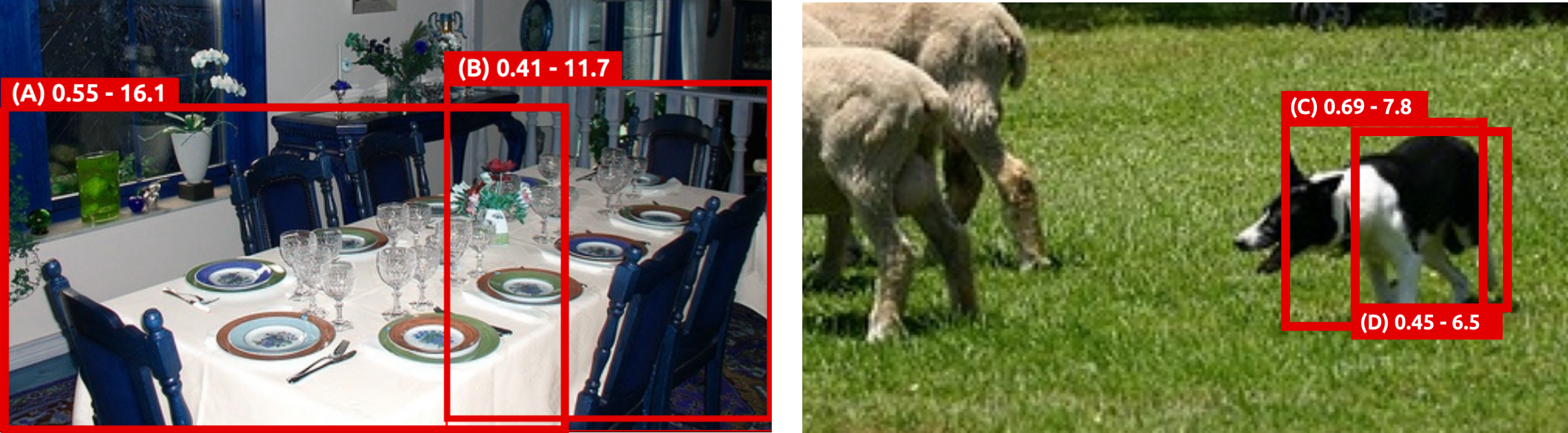}
    \vspace{0.6em}
    \caption{Four examples of region proposals after training only the first VOC10+10 task. Values shown are IOUs with the ground truth bounding box of the table and the dog, and the maximal logit of the ROI-head prediction of that proposal.}
    \label{fig:proposal_examples}
\end{figure}

During training on earlier tasks, RPNs already detect objects of future tasks even though they aren't labeled yet (see Supplementary). As a result, the teacher will learn to confidently classify these objects as earlier classes or background, which leads to conflicting losses when combining distillation with the cross-entropy classification loss of the ROI-heads. To illustrate this effect, we show  four high scoring region proposals of the RPN after the first task (T1) of VOC10+10 in Figure~\ref{fig:proposal_examples}. In Faster ILOD, all four are candidates to use in the distillation loss. We claim that these wrongly classified proposals hinder the learning of the T2 classes, since they overlap more with a T2 object (\emph{e.g.}~table and dog) than one from T1 (\emph{e.g.}~chair). Next, we uncover the issues caused by these proposals further and propose two techniques to alleviate them.

\subsubsection{Selective Distillation}
Ideally, only proposals that contain old objects are used in the distillation of the ROI-heads. To filter out those that do contain new objects, the IOU of the proposed regions with the current ground truths offers an important clue. If they are above a threshold (we use 0.5), it is likely that these regions primarily contain a new object. Because the teacher has never seen the correct label for these, it can not possibly classify these accurately, and its classification should not be trusted. Therefore, before applying distillation, all proposals with IOU higher than 0.5 with any ground truth are filtered out, and only the others are used for distillation. In Figure~\ref{fig:proposal_examples}, this would preclude proposals $A$ and $C$ from being used in the distillation, but not $B$ and $D$.


\subsubsection{Huber Loss}
\label{sec:improve:huber}
\begin{table}[]
\centering
\resizebox{0.75\linewidth}{!}{%
\begin{tabular}{@{}r|llllllllll@{}}
\toprule
& table & dog & horse & motor & person & plant & sheep & sofa & train & tv      \\ 
\midrule
chair co-occurrence & 0.7 & 0.0 & 0.0 & 0.0 & 1.4 & 0.3 & 0.0 & 0.2 & 0.0 & 0.2 \\ 
median bkg. score & 7.4 & 0.9 & 1.8 & 2.9 & 7.0 & 7.8 & 1.5 & 4.6 & 3.3 & 5.1   \\ 
\bottomrule
\end{tabular}%
}
\vspace{0.6em}
\caption{Average number of objects in T2 that are in an image with a chair (T1 object) and the median background score of their ground truth bounding box after T1}
\label{tab:chair_cooc}
\end{table}

The IOU of proposals $B$ and $D$ in Figure \ref{fig:proposal_examples} with any old object's ground truth bounding box is below $0.5$, but they still contain a significant portion of a new object. Ideally, these should not be used for distillation, but lowering the threshold would mean throwing away too many correct proposals with objects of old tasks. This is problematic, since proposals with partly new objects can still lead to conflicting loss signals. In such cases, the increase in distillation loss should be similar to the decrease in CE-loss. Unfortunately, this is not the case: when using MSE, the distillation loss can become large, especially for objects that co-occur often with objects of the previous task. Since the RPN already detected these co-occurring objects during the first task, the model has learned to classify them confidently as background or as a wrong class, see Table \ref{tab:chair_cooc} and the Supplementary. This leads to higher background scores for objects that were often in the past task, compared to those that co-occur less often with past objects. An exception to this observation is the person class, which co-occurs often with the chair and has a high background score, yet still scores well. This is most likely due to the over-representation of this class, which constitutes 60\% of the new task objects. 


\paragraph{}
The higher initial logits of these objects mean their logits have to change more in order to get a correct classification, which eventually leads to a problem with the MSE-loss. If the distillation loss term of proposals like $B$ and $D$ increases faster than the decrease in classification loss of proposals like $A$ and $C$, the regularization will dominate. Since MSE is quadratic, this gets worse the higher the initial logits are. If the $L_1$ distance between the closest correct logits $y_n$ and the initial ones $y_o$ does not grow too large, the MSE is still of the same order of magnitude as the CE-loss. However, for proposals like $D$ this distance is much larger, leading to MSE-losses that blow up and over-regularization. 


\paragraph{}
To alleviate this issue, we propose to change the MSE-loss to a Huber loss for the distillation of the ROI-heads, defined as:
\begin{equation}
\mathcal{L}_h(x, y) = \left\{
        \begin{array}{ll}
            \frac{1}{2}(x-y)^2 \qquad &\text{if } \lvert x - y \rvert < \delta \\
            \delta (\lvert x - y \rvert - \frac{1}{2}\delta) \qquad &\text{otherwise}
        \end{array}
    \right.
\end{equation}
This is quadratic for differences less than $\delta$ and linear for values greater than or equal to $\delta$. For larger differences between correct and initial logits, this ensures that the increase in distillation loss remains of the same order of magnitude as the decrease in CE-loss. In contrast, there isn't a large difference between MSE- and Huber loss for the distillation of old objects, because the difference of their logits will be smaller given that there is no conflicting CE-loss. This leads to the sought after trade-off: more modest distillation for proposals that are close to new objects and require changes in their classification, while remaining strict for other proposals. Adapting MSE losses to a Huber loss has been shown effective in other settings too, see for instance \cite{liu2020incdet}.

\section{Results}
\label{sec:results}

\subsection{Benchmarks and evaluation}
We evaluate our method on three two-task scenario's of VOC2007 \cite{pascal-voc-2007} ($10+10$, $15+5$ and $19+1$) and a $40+40$ scenario in MS COCO \cite{lin2014microsoft}. We primarily compare our improvements to the original version of Faster-ILOD \cite{peng2020faster}, showing how these fixes improve the distillation in Faster-ILOD, which is our main objective. For context, we then provide SOTA in COD that uses more complex mechanisms; Open World Detection (ORE) \cite{joseph2021towards} and Meta \cite{kj2021incremental}. Both of these methods use a replay memory, with respectively 50 and 10 images per class. In Continual Learning, replay and regularization (distillation) are considered complementing ideas~\cite{delange2021continual}, and studying the impact of rehearsal, and its interaction with distillation in COD constitutes future work. The results for Faster-ILOD, ORE and Meta are from \cite{kj2021incremental}, and publicly available T1 models of each method are used to allow a fair comparison. We show the mAP results for each task (T1 and T2), the mAP over all classes and the average of both tasks. Reporting only the final mAP over all classes may hide differences across methods and tasks, especially in the 19+1 benchmark where the single new class does not influence the overall mAP much. For VOC we report the mAP at IOU 0.5, for MS COCO the average mAP at IOUs from 0.5 to 0.95 with step-size 0.05, which both are the respective defaults.

\subsection{Implementation details}
We use a standard Faster RCNN model, with ResNet-50 as backbone, which is the same setting as in the methods we compare to. For all benchmarks, we use SGD with momentum $0.9$. In the 10+10 benchmark, the new task is trained for $20k$ iterations, batch size 4 and learning rate $0.001$, decaying after $15k$ and $17.5k$ iterations. 15+5 is trained for $10k$ iterations, batch size 4, learning rate $0.001$, decaying at iteration $7.5k$. The 19+1 setting is trained with batch size 1, $10k$ iterations and learning rate $0.0001$. The MS Coco benchmark is trained for $90k$ iterations, 16 images per batch and initial learning rate 0.02. These settings were chosen to be similar to both \cite{peng2020faster} and \cite{kj2021incremental}, see Supplementary for a full comparison. The implementation is written in the Detectron2 framework \cite{wu2019detectron2}, and is publicly available. 

\subsection{Results and Analysis}
Tables \ref{tab:our_results} and \ref{tab:coco results} show the results for VOC and COCO. Our adaptation consistently improves over the original Faster-ILOD on all three benchmarks, with an increase of 3.4, 1.6 and 3.8 mAP for the 10+10, 15+5 and 19+1 setting respectively. For the latter two, our adaptation even does better than both SOTA methods, without relying on a rehearsal memory. Compared to Faster-ILOD our method is able to adapt better to the new task, without compromising much on the old task. The same is true for the results on COCO, which improves over Meta and Faster-ILOD. To verify that our adaptations resolved the issues we exposed, we included \textit{Ours} in Figures \ref{fig:roi_analysis} and \ref{fig:roi_out}. The mAPs of the classes that weren't properly learned at first are markedly higher than in the original Faster ILOD. The number of classification mistakes on the second task nearly halved, while only compromising 2\% in the mistakes of the first task. Table \ref{tab:ablation} shows an ablation of the VOC10+10 task, which shows that both the Selective Distillation and the Huber loss have a positive influence, and combining both leads to the best results.

\begin{table}[t]
\centering
\resizebox{1.0\textwidth}{!}{%
\begin{tabular}{@{}rccccc|ccccc|ccccc@{}}
\hline
\multicolumn{1}{c}{} & \multicolumn{5}{c|}{10+10}                                    & \multicolumn{5}{c|}{15+5}                            & \multicolumn{5}{c}{19+1}                             \\ 
\multicolumn{1}{c}{} & First 10 & FI  & Ours                 & ORE  & Meta          & First 15 & FI   & Ours                 & ORE  & Meta & First 19 & FI   & Ours                 & ORE  & Meta \\ \hline
T1                   & 68.7     & 69.8 & 66.2 (-3.6)          & 60.4 & 68.3          & 74.2     & 71.6 & 73.3 (+1.7)          & 71.8 & 71.7 & 73.2     & 68.9 & 72.8 (+3.9)          & 67.9 & 70.9 \\
T2                   & -        & 54.5 & 64.7 (+10.2)         & 68.8 & 64.3          & -        & 56.9 & 58.2 (+1.3)          & 58.7 & 55.9 & -        & 61.1 & 62.8 (+1.7)          & 60.1 & 57.6 \\
Task Average         & -        & 62.1 & \textbf{65.5 (+3.4)} & 64.6 & \textbf{66.3} & -        & 64.3 & \textbf{65.7 (+1.4)} & 65.2 & 63.8 & -        & 65.0 & \textbf{67.8 (+2.8)} & 64.0 & 64.2 \\
mAP                  & -        & 62.1 & \textbf{65.5 (+3.4)} & 64.6 & \textbf{66.3} & -        & 67.9 & \textbf{69.5 (+1.6)} & 68.5 & 67.8 & -        & 68.5 & \textbf{72.3 (+3.8)} & 67.5 & 70.2 \\ \hline
\end{tabular}%
}
\vspace{0.6em}
\caption{Results on three two-task VOC benchmarks, 10+10, 15+5 and 19+1. The results are VOC mAP @IOU 0.5 and tested on the full test set, as explained in Section \ref{sec:method:analysis}. We compare ours to Faster-ILOD \cite{peng2020faster} (FI), the values in brackets show the difference. Additionally, we compare to SOTA methods ORE \cite{joseph2021towards} and Meta \cite{kj2021incremental}, which both use a replay memory. The initial mAP is in the column first X. For each method the mAP of the first (T1) and second (T2) task is shown, as well as the overall mAP and the average mAP of both tasks}
\label{tab:our_results}

\end{table}

\begin{table}[t]
    \centering
    \resizebox{0.30\textwidth}{!}{%
    \begin{tabular}{@{}r|lll@{}}
        \toprule
                    & AP   & AP$_{50}$ & AP$_{75}$ \\ \midrule
        Joint       & 31.5 & 50.9       & 33.5       \\ \midrule
        Faster-ILOD & 20.6 & 40.1       & -          \\
        Meta        & 23.8 & 40.5       & 24.4       \\
        Ours        & \textbf{26.4} & \textbf{44.3}       & \textbf{27.4}       \\ \bottomrule
    \end{tabular}
    } %
    \qquad 
    \resizebox{0.40\textwidth}{!}{%
    \begin{tabular}{@{}r|cccc@{}}
    \toprule
              & T1 & T2 & mAP \\ \midrule
        Faster ILOD   & 69.8 & 54.5 & 62.1   \\
        Huber         & \textbf{71.2} & 58.0 & 64.6   \\
        Selective Distillation & 69.5 & 59.7 & 64.6   \\
        Both          & 66.2 & \textbf{64.7} & \textbf{65.5}   \\ \bottomrule
    \end{tabular}
    } %
    \vspace{0.6em}
    \caption{\textit{(Left)} AP, AP50 and AP75 on the minival of MS COCO40+40. AP is the average of the APs at IOU levels from 0.5 to 0.95 with steps of 0.05. Results on the full validation set and per task are in the Supplementary.}
    \label{tab:coco results}
    \vspace{0.6em}
    \caption{\textit{(Right)} Ablation study of the VOC10+10 task with Selective Distillation and/or Huberloss. Additional ablations are available in the Supplementary.}
    \label{tab:ablation}    
\end{table}

\section{Limitations}
We identify a number of limitations in our study that can lead to further investigation. Namely; all three datasets used include images with annotations for \emph{both} tasks, albeit unlabeled. Absence of such images may lead to different results. 
Furthermore, we have endeavored to account for all mechanisms that contribute to both the aspects of (1) forgetting; and (2) prevention of new-task learning, however the inherent complexity of object detectors dictates that additional causal factors may yet be uncovered. Finally we note that our class-incremental experimental work currently considers only two-task benchmarks. Extensions to longer task sequences will enable insight into distillation mechanisms in such additional challenging scenarios.

\section{Conclusion}
We undertook a thorough analysis of how models in COD forget and how distillation methods, such as Faster-ILOD, limit forgetting. We identified issues that prevent Faster-ILOD from learning new tasks and propose solutions to fix these problems. Our modifications provide improvement to the current SOTA, on multiple datasets and benchmarks. 

\section{Acknowledgment}
We gratefully acknowledge the support of MindSpore, CANN (Compute Architecture for Neural Networks) and Ascend AI Processor used for this research.

\clearpage
\newpage
\bibliography{egbib}

\appendix
\include{appendix}

\end{document}

%% file: appendix.tex
\section{Method and Analysis}
In this section we include extra details on the analysis of Faster-ILOD and our improvement. We show the confusions matrices for both methods, the co-occurences of all objects in VOC, and an elaborate discussion on the differences between Huber and MSE-loss.

\subsection{Confusion Matrices VOC10+10}
\begin{figure}[h]
    \centering
    \includegraphics[width=1.0\linewidth]{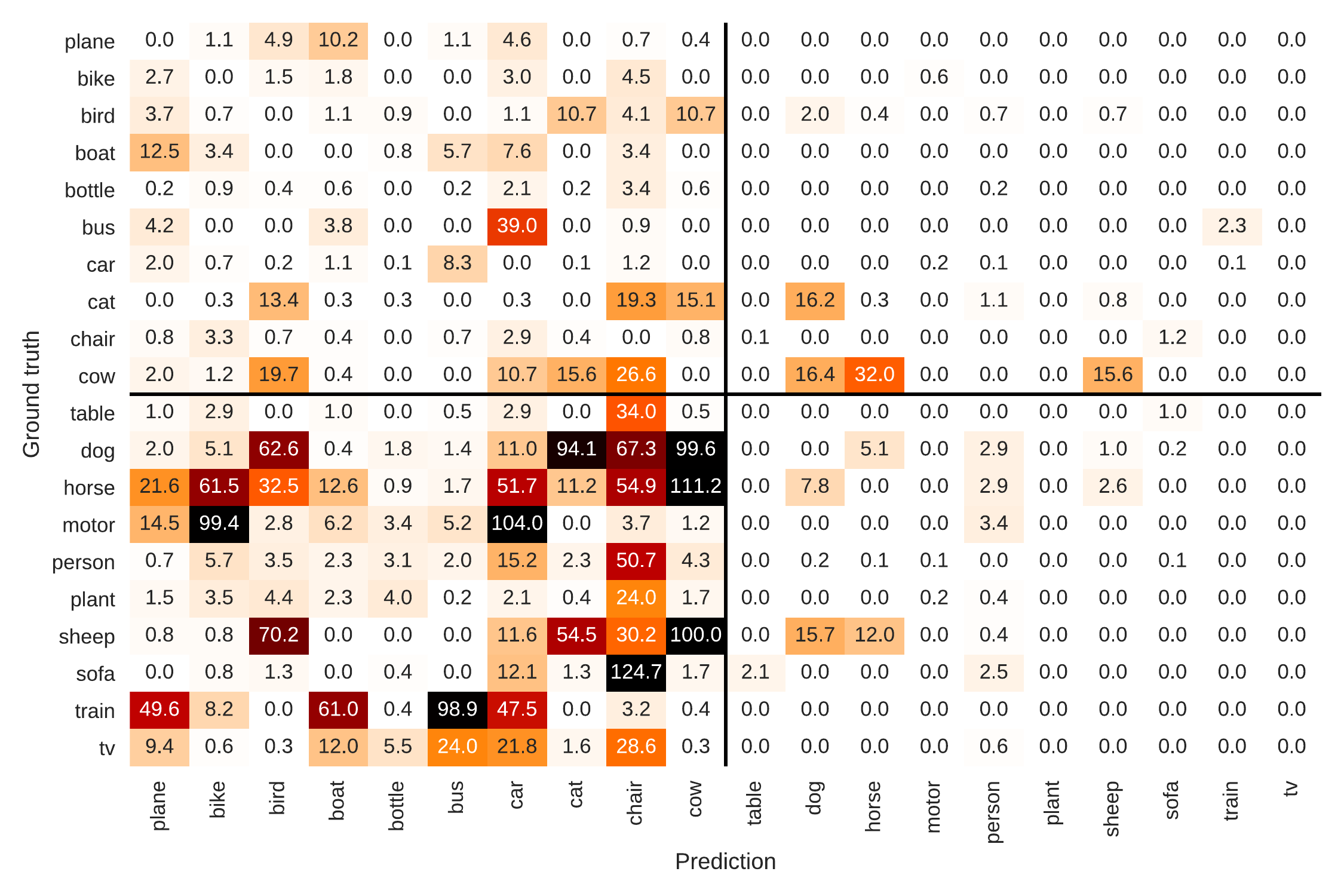}
    \caption{Confusion matrix showing wrongly classified detections in the VOC10+10 task, trained with Faster-ILOD. The results are normalized by the number of ground truth bounding boxes of each class.}
    \label{fig:full_confusion_filod}
\end{figure}

\begin{figure}[h]
    \centering
    \includegraphics[width=1.0\linewidth]{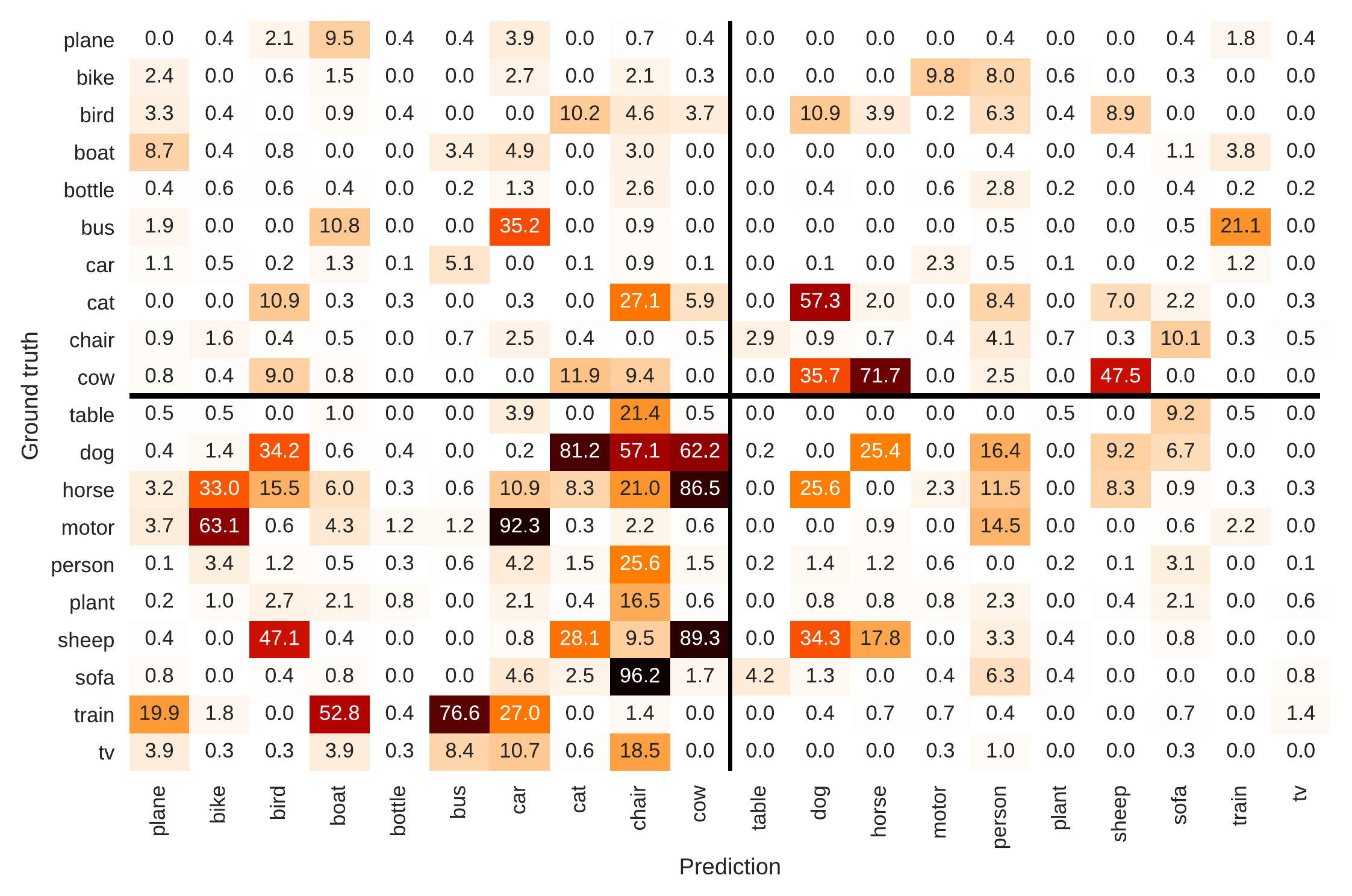}
    \caption{Confusion matrix showing wrongly classified detections in the VOC10+10 task, trained with our method. The results are normalized by the number of ground truth bounding boxes of each class.}
    \label{fig:full_confusion}
\end{figure}

Figure \ref{fig:full_confusion_filod} shows the confusion matrices after learning both tasks in the VOC10+10 benchmark, with Faster-ILOD, and Figure \ref{fig:full_confusion} shows that of our method. The rows are normalized by the number of ground truths for each class. It is possible to have multiple wrongly classified proposals for a single ground truth, which means that the rows do not sum to $100\%$. $84\%$ of \textit{all} wrongly classified proposals in the Faster-ILOD model are T2 objects that are classified as T1 objects (lower-left corner), indicating the model is too stable. In our method this has dropped to $65\%$, while also having $26\%$ less wrongly classified proposals in total. In absolute numbers, Faster-ILOD has $10.159$ T2 proposals classified as T1 objects, while ours only has $5641$. 

\subsection{Co-occurrence matrices in VOC}
\begin{figure}[h]
    \centering
    \includegraphics[width=1.0\linewidth]{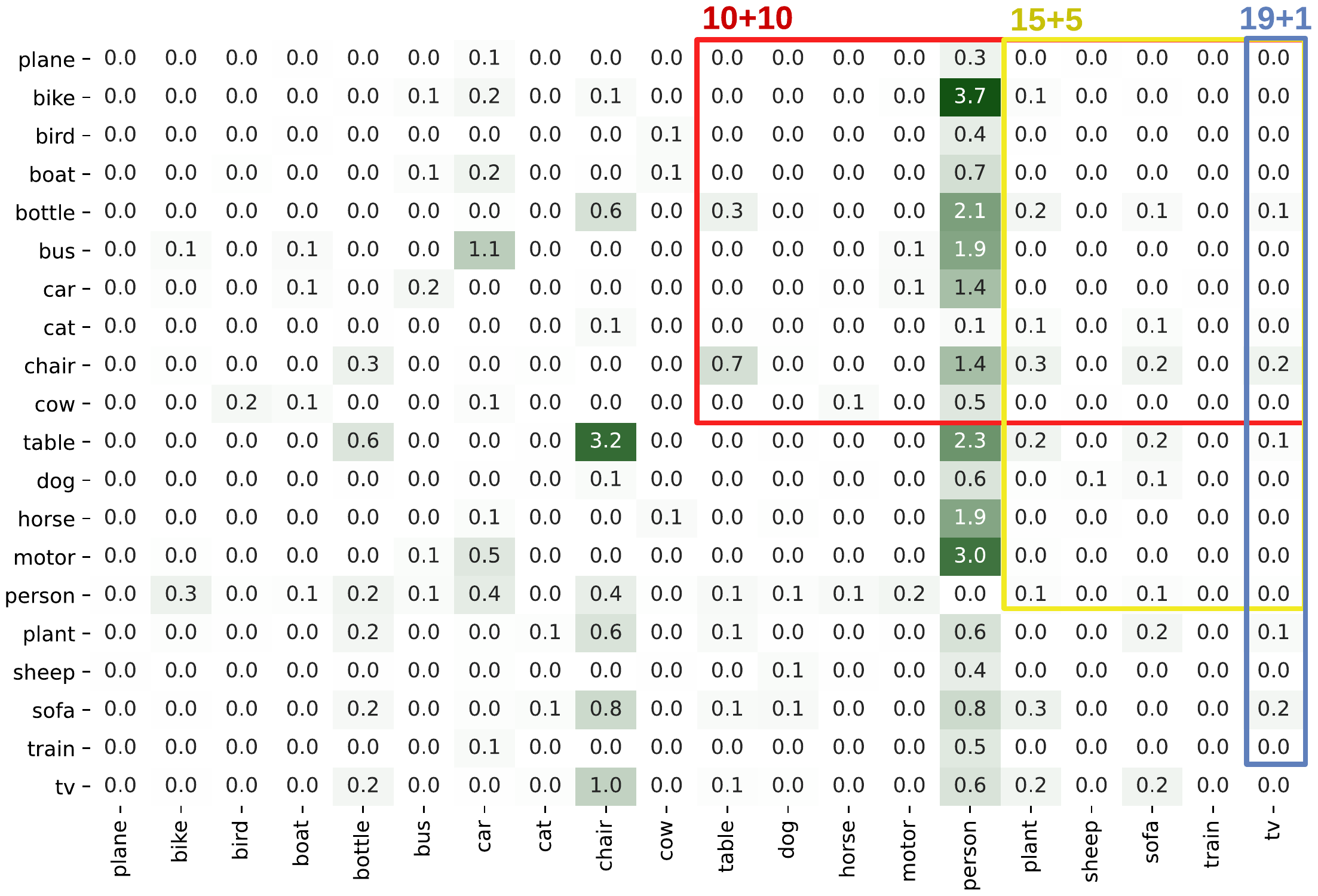}
    \caption{Co-occurency matrix for the classes in VOC. Values indicate the average number of `column' objects that appear together in images with a `row' object. \emph{E.g.} for each chair, there is on average $0.7$ tables in an image, while for each table there are on average $3.2$ chairs.}
    \label{fig:full_cooc}
\end{figure}

Figure~\ref{fig:full_cooc} provides the full co-occurency matrix of the VOC2007 train and validation objects. Each row represents the average number of the other classes in images with that object. \emph{E.g.}, in each image with a plane, there is on average 0.1 cars and 0.3 persons. With $35\%$ of all objects, it is not surprising that the person object stands out and occurs often in images with other objects. However, the most important part of this matrix is the upper right sub-matrix ($10 \times 10$, $15 \times 5$, $19 \times 1$, for respective benchmarks) since these show how often a new object has appeared in images of previous tasks. For those images, both the distillation and classification loss are important, see Section \ref{sec:improving_distillation} for further details.

\subsection{Additional Details on Competing Losses}

\begin{figure}[tb]
    \centering
    \includegraphics[width=0.6\linewidth]{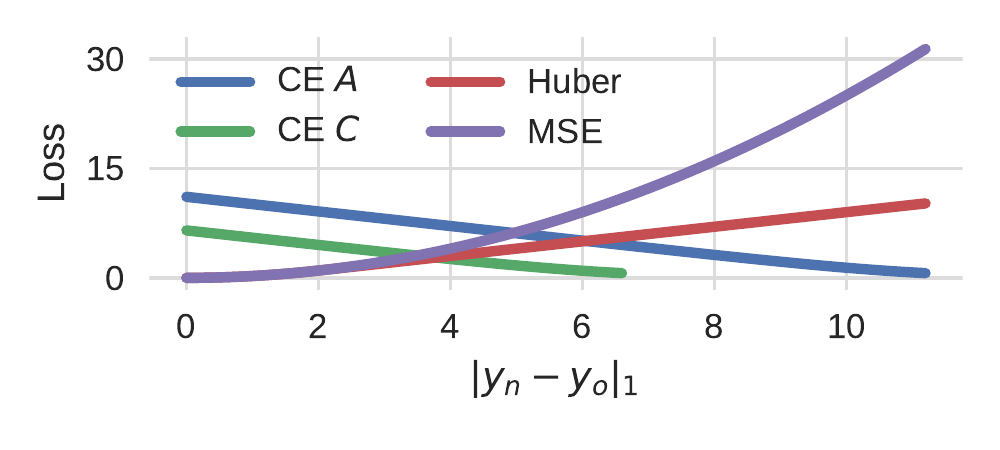}
    \caption{Different loss functions when the prediction changes from a wrong set of logits $y_o$ to a correct set $y_n$. $A$ and $C$ refer to the example proposals in Figure \ref{fig:proposal_examples}, of which the initial logits are used as an example. For the purpose of this illustration we chose $y_n$ such that the $L_2$ distance to $y_o$ is minimal, which results in the smallest MSE-loss.}
    \label{fig:mse_huber_example}
\end{figure}

In Figure \ref{fig:mse_huber_example} and Section \ref{sec:improve:huber}, we showed that for large differences between initial and correct logits the MSE loss can become proportionally much larger than the corresponding CE-loss. Here, we elaborate a bit further on the example given in the main paper.

\paragraph{}
Our illustration assumed a binary classification model. At the start of the new task, the output of the network is $x_o = (l, 0)$. In the illustration we set $l$ equal to the maximal logits of the teacher network for proposals $A$ and $C$ in Figure \ref{fig:proposal_examples}. Given these initial predictions, the set of logits that classifies correctly with the lowest MSE-loss is $x_n = (\frac{l}{2} - \epsilon, \frac{l}{2} + \epsilon)$, with $\epsilon \rightarrow 0$. Points on the $x-$axis in Figure \ref{fig:mse_huber_example} are the $L_1$ distance between the  points of the linear interpolation of $x_n$ and $x_o$ ($x = \alpha x_o + (1 - \alpha) x_n$, $\alpha \in (0, 1)$) and $x_o$ itself. The losses are calculated for each set of logits $x$, with the distillation losses calculated relative to the initial set $x_o$. Realistic models evidently have more than two logits. Assuming additional logits (that neither represent the wrong nor the correct class) are added, and given that they don't change, the distillation losses will stay equal, and only the CE-loss will decrease. CE-loss is given by

\begin{equation}
    \mathcal{L}_\text{CE} = \frac{e^{y_c}}{\sum_i e^{y_i}}
\end{equation}

with $y$ the logits yielded by the network, and $y_c$ the correct class. Adding more, incorrect, logits only changes the denominator and since the exponential function is always positive, $\mathcal{L}_\text{CE}$ can only decrease. Of course, it is possible that these extra logits change too during training of the new task. Yet, relative to the change in the wrong and correct logits these changes are small. 

\subsection{Backbone and RPN experiments}
\paragraph{}
The backbone experiment of \ref{sec:backbone_rpn} is straight-forward. First, we train the model on T1 of VOC10+10. Then, we freeze the backbone and train the RPN and classification heads as in Faster-ILOD. The second experiment does the same thing, but rather than freezing the backbone it is fine-tuned without any additional distillation losses. Neither of these experiments lead to significant change in results, suggesting that forgetting in these tasks isn't primarily caused by a shift in the backbone's output distribution.

\begin{figure*}[tb]
\centering
\begin{subfigure}[t]{0.45\textwidth}
    \includegraphics[width=\textwidth]{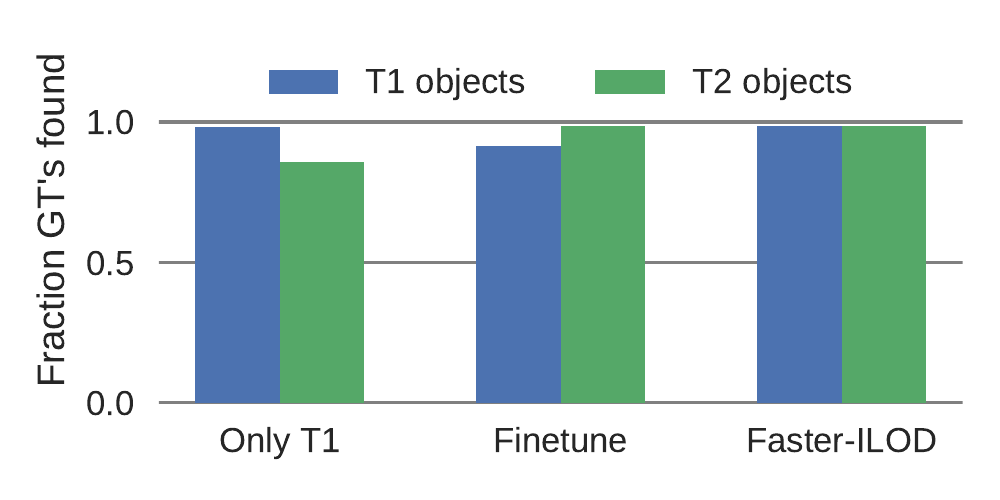}
\end{subfigure}
\begin{subfigure}[t]{0.45\textwidth}
    \includegraphics[width=\textwidth]{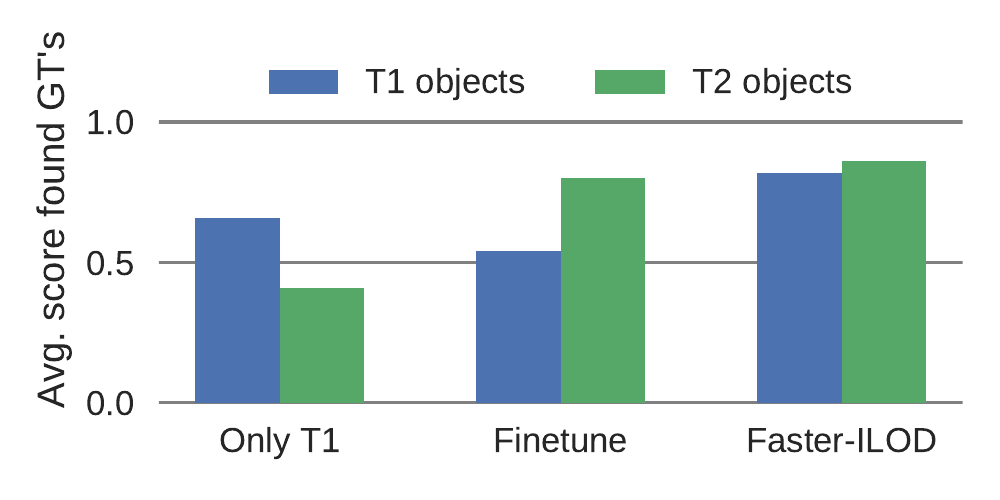}
\end{subfigure}
\caption{Fraction of objects found and their average objectness score for a model trained only on T1 objects, one that's subsequently also finetuned on T2 and one that uses distillation like Faster-ILOD.}
\label{fig:rpn_anlysis}
\end{figure*}

\paragraph{}
The RPN can fail to detect an object either by missing it entirely, or by assigning it an objectness score that is too low. To evaluate if and how the RPN forgets, we record all ground truth bounding boxes that are detected by the RPN. Following the standard object detection criterion, those are the RPN detections obtained during inference with an IOU larger than 0.5 with ground truth annotations. In Figure~\ref{fig:rpn_anlysis}, we compare the percentage of found ground truths annotations and their average objectness score, for a model that is only trained on the 10 classes of the first task, one that is subsequently finetuned on the second task, and a Faster-ILOD distillation based model. Even after only having seen objects of T1, the RPN finds 85.6\% of all objects of T2, albeit with a lower score. Finetuning on T2 improves this to near perfection, at the cost of a small drop in detected T1 ground truth labels (-5.8\%) and a lower score. Using MSE to regularize the RPN (Faster-ILOD), it finds nearly all objects of \textit{both} T1 and T2 at IOU 0.5, while having comparable scores for both tasks.

\section{Extra Results}
Here, we make a comparison of the hyper-parameters we used and those used in the methods we compare to. We add results on the full validations set of MS COCO and additional ablations using only Huber Loss or Selective Distillation. Finally, we include the per class APs of the $10+10$, $15+5$ and $19+1$ VOC benchmarks. 

\subsection{Hyperparameters}
We based the hyper-parameters for the training of our models based on the implementations of Faster-ILOD and Meta, see Table \ref{tab:hyper-table} for an overview of all values. Meta always used 16 images per batch, which was computationally infeasible for the hardware available to us (except for MS COCO, for which we relied on other machines). The VOC benchmarks were trained on machines using a single NVIDIA GeFore RTX 3090 or similar. Training of the second VOC tasks in 10+10, 15+5 and 19+1 took respectively around 9, 8 and 5 hours of GPU-time. 

\begin{table}[h]
\centering
\begin{tabular}{@{}ccc|cccc@{}}
\toprule
Dataset               & Benchmark              & Method      & Lr    & Batch Size & Total Iter. & Lr. Schedule \\ \midrule
\multirow{9}{*}{VOC}  & \multirow{3}{*}{10+10} & Faster-ILOD & 0.001 & 1          & 40k         & 30k          \\
                      &                        & Meta        & 0.02  & 16         & 18k + 3k    & 10k, 14k     \\
                      &                        & Ours        & 0.001 & 4          & 20k         & 15k, 17.5k   \\ \cmidrule{2-7}
                      & \multirow{3}{*}{15+5}  & Faster-ILOD & 0.001 & 1          & 40k         & 30k          \\ 
                      &                        & Meta        & 0.02  & 16         & 9k + 3k     & ?            \\
                      &                        & Ours        & 0.001 & 4          & 10k         & 7.5k         \\ \cmidrule{2-7}
                      & \multirow{3}{*}{19+1}  & Faster-ILOD & 1e-4  & 1          & 5k          & /            \\
                      &                        & Meta        & 0.02  & 16         & 5k + 3k     & ?            \\
                      &                        & Ours        & 0.001 & 1          & 10k         & 7.5k         \\ \cmidrule{2-7}
\multirow{3}{*}{COCO} & \multirow{3}{*}{40+40} & Faster-ILOD & 1e-4  & 1          & 400k        & ?            \\
                      &                        & Meta        & 0.02  & 16         & 90k         & 60k, 80k     \\
                      &                        & Ours        & 0.02  & 16         & 90k         & 60k, 80k     \\ \bottomrule
\end{tabular}
\vspace{1.5em}
\caption{Hyper-parameters used in Faster-ILOD \cite{peng2020faster} and Meta \cite{kj2021incremental}, on which \textit{Ours} is based. For Meta the additional $3k$ iterations refers to their finetuning step. The learning rate schedule are multi-step schedulers, and show at which iteration the learning rate gets divided by $10$. `?' indicates the value wasn't clear from their code or paper}
\label{tab:hyper-table}
\end{table}

\subsection{Additional Results on the CLAD-D Domain Incremental Benchmark}
\begin{figure}[h]
\centering
\begin{subfigure}[t]{0.24\textwidth}
    \includegraphics[width=\textwidth]{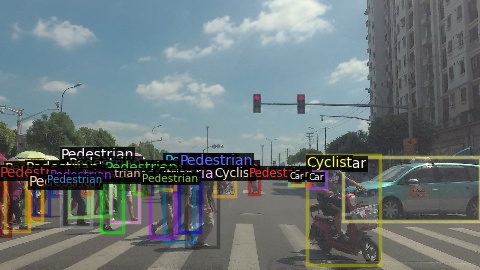}
    \caption{Task 1: Clear weather, at daytime, in the city center. 4470/500 images.}
\end{subfigure}
\begin{subfigure}[t]{0.24\textwidth}
    \includegraphics[width=\textwidth]{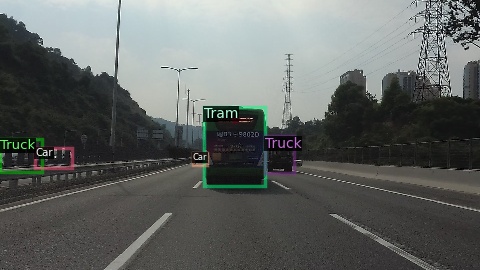}
    \caption{Task 2: Daytime images on the highway. 1329/148 images.}
\end{subfigure}
\begin{subfigure}[t]{0.24\textwidth}
    \includegraphics[width=\textwidth]{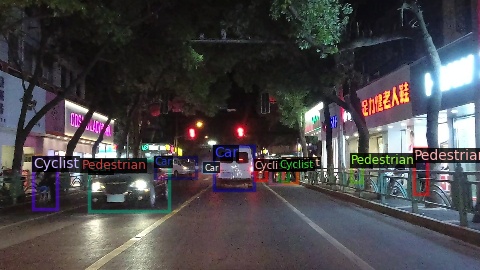}
    \caption{Task 3: Images at night. 1480/166 images.}
\end{subfigure}
\begin{subfigure}[t]{0.24\textwidth}
    \includegraphics[width=\textwidth]{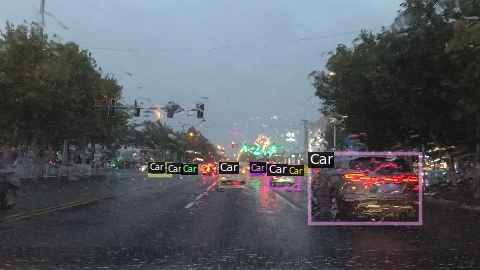}
    \caption{Task 4: Rainy images. 524/59 images.}
\end{subfigure}
\vspace{0.5em}
\caption{Example images and number of images in train/test set, for the four tasks in the domain incremental benchmark of SODA 10M.}
\label{fig:soda_10m}
\end{figure}

\begin{figure}[h]
    \centering
    \includegraphics[width=0.65\linewidth]{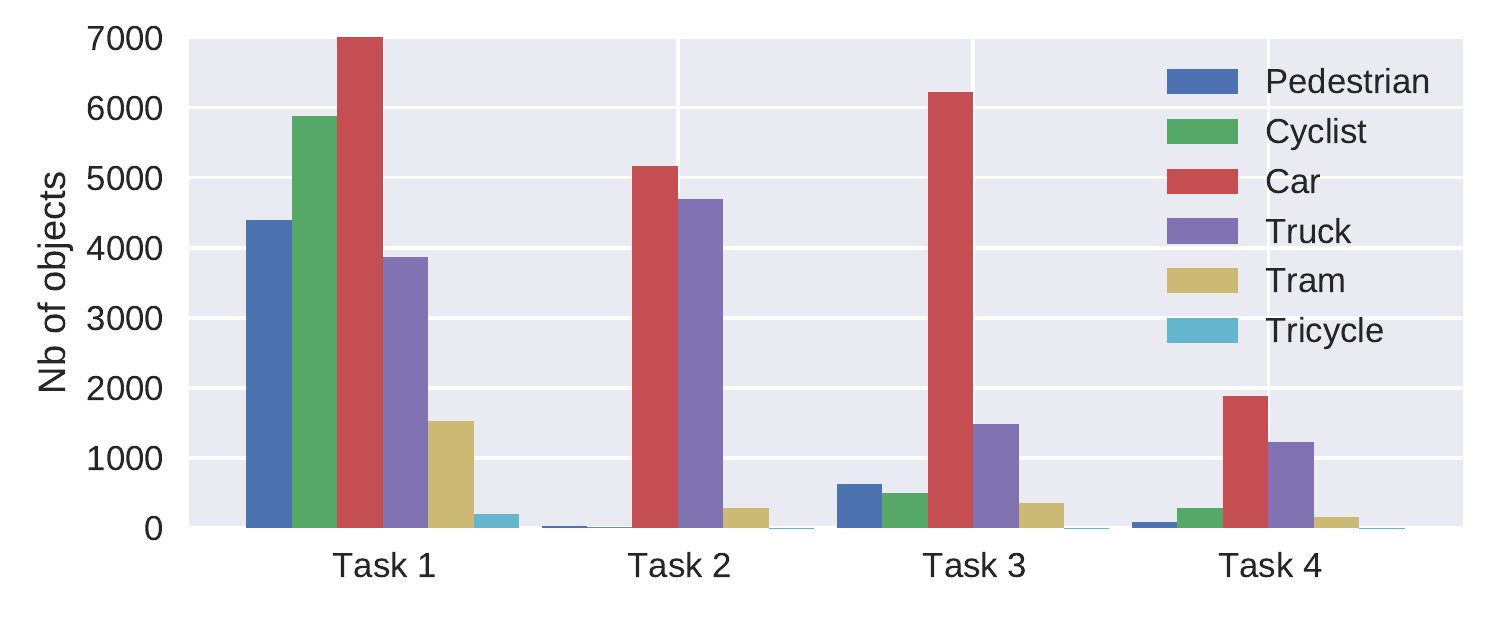}
    \caption{Number of objects for each category in the SODA 10M benchmark. In the first task, there are actually $21,156$ cars, but for the sake of visibility, the graph is truncated. The number of objects per class in the test set is similarly distributed.}
    \label{fig:soda_10m_categories}
\end{figure}

The CLAD-D benchmark of the SODA10M challenge at ICCV2021 is centered around domain incremental continual learning. In four tasks, four domains are covered: (1) clear weather, at daytime, in the city center, (2) daytime images on the highway, (3) images at night and (4) rainy images. Each of these domains has different characteristics, requiring to model to be inert to these changes in the input distribution of the images (\emph{e.g.} a car looks different at night then during the day, the perspective of other cars is different on the highway than in the city center), see Figure \ref{fig:soda_10m} for representative examples for each domain. With domain incremental challenges come weak class incremental challenges: there is a large imbalance in the appearance of certain categories in some domains. Pedestrians and cyclists are only rarely seen on highways (luckily), but this makes the training model susceptible to catastrophic forgetting of said categories. See Figure \ref{fig:soda_10m_categories} for an overview of the number of categories per domain. The exact splits and pre-processing of the images are the same as those of the ICCV challenge, so we refer to their code base for further details \cite{challenge}.

\paragraph{}
We firstly evaluate the fine-tuning baseline in Table~\ref{tab:di_results}. It may be observed that catastrophic forgetting is still a problem in the domain-incremental setting. The mAPs of Task 1 - 3 drop by $11.2$, $4.1$, and $13.2$, respectively, after all tasks are learned. This shows that, albeit less catastrophically, the domain incremental setting also suffers from forgetting. Faster-ILOD~\cite{peng2020faster} significantly reduces the forgetting on task 1 and 3. However, this stability comes at the cost of losing plasticity. Comparing the results of a task just after it has been learned (the entries on the main diagonal in Table~\ref{tab:di_results}) for fine-tuning and Faster-ILOD, we can see that Faster-ILOD's student has difficulty learning new tasks, with mAPs that are lower than fine-tuning. Because of the lack of plasticity in Faster-ILOD, the final model is worse than the final fine-tuning model, by $4.2$ mAP. Our method also improves Faster-ILOD on this benchmark, by $3.7$ mAP points after the fourth task. Yet, it falls just short of fine-tuning by $0.5$ mAP. However, the results for fine-tuning are heavily influenced by the previous task, and the  advantage is lost as soon as a new task is learned. In contrast, the performance of our method is more balanced. Notably, it is able to generate scores broadly competitive with ORE~\cite{joseph2021towards}, which relies explicitly on replay memory of at least $50$ samples per task, \emph{vs.} our approach (no replay).

\begin{table}[t]
\centering
\resizebox{\textwidth}{!}{%
\begin{tabular}{@{}lccccc|ccccc|ccccc|ccccc@{}}
\toprule
           & \multicolumn{5}{c|}{Fine-tuning}  & \multicolumn{5}{c|}{Faster-ILOD \cite{peng2020faster}}      
           & \multicolumn{5}{c|}{Ours}      & \multicolumn{5}{c}{ORE \cite{joseph2021towards}}      \\
           & T1   & T2   & T3   & T4   & Task Avg. & T1   & T2   & T3   & T4   & Task Avg. 
           & T1   & T2   & T3   & T4   & Task Avg. & T1   & T2   & T3   & T4   & Task Avg.     \\ \midrule
After T1   & 64.2 & -    & -    & -    & 64.2                 & 64.2 & -    & -    & -    & 64.2      
           & 64.2 & -    & -    & -    & 64.2 (+0.0)          & 63.7 & -    & -    & -    & 63.7          \\
After T2   & 53.5 & 54.3 & -    & -    & 53.9                 & 60.1 & 49.5 & -    & -    & 54.8      
           & 62.9 & 50.8 & -    & -    & 56.8 (+2.0) & 58.3 & 59.8 & -    & -    & \textbf{59.0}          \\
After T3   & 51.2 & 44.2 & 74.7 & -    & 56.7                 & 59.3 & 43.4 & 55.6 & -    & 52.8      
           & 61.6 & 47.1 & 64.4 & -    & \textbf{57.7 (+4.9)} & 56.9 & 44.4 & 72.1 & -    & \textbf{57.8}          \\
After T4   & 53.0 & 50.1 & 61.5 & 74.4 & 59.8                 & 59.3 & 43.9 & 55.4 & 63.7 & 55.6
           & 60.7 & 49.3 & 63.2 & 64.1 & 59.3 (+3.7)          & 57.3 & 52.0 & 66.9 & 64.3 & \textbf{60.1} \\ \midrule
Forgetting & 11.2  & 4.1 & 13.2 & -    & 9.5                  & 4.9  & 5.6  & 0.2  & -    & 3.6       
           & 3.5   & 1.5 & 1.1  & -    & \textbf{2.0}         & 6.4  & 7.8  & 5.2  & -    & 6.5           \\ \bottomrule
\end{tabular}%
}
\vspace{1.5em}
\caption{Fine-tuning, Faster-ILOD, ORE and our results on the CLAD-D domain incremental benchmark. For each task the mAP @IOU0.5 is shown. Values in a column are the mAP of that task, evaluated after each task. The forgetting shown is between the first time a task is evaluated and the last time. The values in brackets in Ours show the difference with Faster-ILOD}
\label{tab:di_results}

\end{table}

\subsection{Additional results on MS COCO}
In addition to the results for MS COCO on the mini-validation set (which are the first 5000 images of the full validation set), here we provide results on the full validation set as well as results per task. Both Faster-ILOD and Meta do not provide per task results on COCO, hence we can only show ours per task. The results on the full-validation set confirm the conclusions of the mini validation set, and our method improves vastly over Meta and Faster-ILOD. The AP for both tasks is comparable, indicating that stability and plasticity are, also in this benchmark, well balanced.

\begin{table}[h]
\centering
\begin{tabular}{@{}l|r|ccc@{}}
\toprule
     eval. set    & Method      & AP T1 & AP T2 & Avg. AP       \\ \midrule
         & Joint       & -     & -     & 31.5          \\
mini-val & Faster-ILOD & -     & -     & 20.6          \\
         & Meta        & -     & -     & 23.8          \\
         & Ours        & 25.0  & 27.8  & \textbf{26.4} \\ \cmidrule(l){1-5} 
         & Joint       & -     & -     & 31.2          \\
full-val & Faster-ILOD & -     & -     & -             \\
         & Meta        & -     & -     & 23.7          \\
         & Ours        & 25.2  & 27.9  & \textbf{26.6} \\ \bottomrule
\end{tabular}
\vspace{1.5em}
\caption{MS COCO results on the mini-validation and the full validation set and per task for our method. Values are standard MS COCO AP, averaged over IOUs from 0.5 to 0.95 with step size 0.05}
\label{tab:more_coco_results}
\end{table}

\subsection{Additional Ablations}
In Table \ref{tab:ablation} we showed that both the Huber loss and Selective Distillation have a positive influence on the VOC10+10 benchmark. In Table \ref{tab:ablation:ci} we complete the ablation with results for VOC 15+5 and 19+1, and in Table \ref{tab:ablation:di} for the domain incremental SODA10M benchmark. These results follow the same trends as those of VOC10+10: individually both the Huber loss and Selective Distillation are helpful, and combining both leads to the best result on all benchmarks.

\begin{table}[h]
\centering
\resizebox{1.0\textwidth}{!}{%
\begin{tabular}{@{}r|cccc|cccc|cccc@{}}
\toprule
& \multicolumn{4}{c|}{10+10}                  & \multicolumn{4}{c|}{15+5}                   & \multicolumn{4}{c}{19+1}                    \\
& T1   & T2   & mAP           & Task Avg.     & T1   & T2   & mAP           & Task Avg.     & T1   & T2   & mAP           & Task Avg.     \\ \midrule
Faster ILOD & 69.8 & 54.5 & 62.1          & 62.1          & 71.6 & 57.0 & 67.9          & 64.3          & 68.9 & 61.1 & 68.5 & 65.0 \\
Huber   & 71.2 & 58.0 & 64.6          & 64.6          & 73.6 & 57.3 & \textbf{69.5} & 65.4          & 72.8 & 59.9 & 72.2       & 66.3          \\
Selective Dist. & 69.5 & 59.7 & 64.6          & 64.6          & 73.6 & 55.6 & 69.1          & 64.6          & 72.9 & 58.2 & 72.2          & 65.6          \\
Both    & 66.2 & 64.7 & \textbf{65.5} & \textbf{65.5} & 73.3 & 58.2 & \textbf{69.5} & \textbf{65.7} & 72.8 & 62.8 & \textbf{72.3} & \textbf{67.8} \\ \bottomrule
\end{tabular}%
}
\vspace{1.5em}
\caption{Individual contribution of the Huber loss and the Selective Distillation, compared to using neither (\emph{e.g.} Faster-ILOD) or both. Results are mAP @IOU 0.5, on the full testsets of benchmarks VOC 10+10, 15+5 and 19+1.}
\label{tab:ablation:ci}
\end{table}

\begin{table}[h]
\centering
\begin{tabular}{@{}r|ccccc@{}}
\toprule
Average after training:  & T1 & T2 & T3 & T4 \\ \midrule
Faster-ILOD & 64.2 & 54.8             & 52.8             & 55.6  \\
Huber    & 64.2 & 54.8             & 56.9             & \textbf{59.6}  \\
Selective Dist.    & 64.2 & 54.8             & 55.6             & 57.2  \\
Both & 64.2 & \textbf{56.8}    & \textbf{57.7}    & \textbf{59.4} \\ \bottomrule
\end{tabular}%
\vspace{1.5em}
\caption{Individual contribution of the Huber loss and the Selective Distillation compared to using neither (\emph{i.e.} Faster-ILOD baseline) or both. Results are mAP @IOU 0.5, on the full testsets for the domain-incremental SODA10M benchmark. Values shown are the average of all tasks learned up to that point (\emph{e.g.} the T3 column is the average of T1-T3, after learning T3)}
\label{tab:ablation:di}
\end{table}

\subsection{Per Class Results VOC}
In Table \ref{tab:results_table}, the individual results per class are shown after learning both tasks in the incremental VOC benchmarks. 

\begin{table}[h]
\centering
\resizebox{\textwidth}{!}{%
\begin{tabular}{@{}r|cccccccccccccccccccc@{}}
\toprule
Benchm. & plane & bike & bird & boat & bottle & bus  & car  & cat  & chair & cow & table & dog  & horse & motor & person & plant & sheep & sofa & train & tv \\ \midrule
10+10    & 74.1 & 79.0 & 67.8 & 55.2 & 58.4 & 69.6 & 84.0 & 75.3 & 43.3 & 55.5 & 52.4 & 72.9 & 81.8 & 75.4 & 76.9 & 31.3 & 66.9 & 58.7 & 67.0 & 64.0 \\
15+5     & 82.7 & 82.6 & 73.3 & 57.3 & 57.5 & 74.8 & 87.2 & 84.1 & 46.0 & 67.4 & 62.3 & 82.6 & 85.0 & 77.4 & 79.3 & 34.9 & 66.6 & 53.8 & 71.0 & 64.6 \\
19+1     & 78.8 & 82.5 & 75.5 & 54.1 & 62.0 & 80.7 & 87.2 & 84.3 & 47.0 & 78.3 & 59.5 & 84.9 & 84.2 & 81.6 & 81.2 & 46.2 & 73.9 & 63.7 & 78.2 & 62.8 \\ \bottomrule
\end{tabular}%
}
\vspace{1.5em}
\caption{AP per class for the VOC benchmarks 10+10, 15+5 and 19+1 for our method. Results are standard VOC AP at IOU $0.5$}
\label{tab:results_table}
\end{table}